\def\year{2020}\relax
\documentclass[letterpaper]{article} % DO NOT CHANGE THIS
\usepackage{aaai20}  % DO NOT CHANGE THIS
\usepackage{times}  % DO NOT CHANGE THIS
\usepackage{helvet} % DO NOT CHANGE THIS
\usepackage{courier}  % DO NOT CHANGE THIS
\usepackage[hyphens]{url}  % DO NOT CHANGE THIS
\usepackage{graphicx} % DO NOT CHANGE THIS
\usepackage{graphicx}  % DO NOT CHANGE THIS
\usepackage{multirow} 
\usepackage{amsmath} 
\usepackage{enumitem}
\usepackage{booktabs}
\usepackage{amsfonts}

\title{Hierarchical Contextualized Representation  for Named Entity Recognition}
\author{
Ying Luo \textsuperscript{\rm 1,2,3},
Fengshun Xiao\textsuperscript{\rm 1,2,3},
Hai Zhao\textsuperscript{\rm 1,2,3,}\thanks{Corresponding author. This paper was partially supported by National Key Research and Development Program of China (No. 2017YFB0304100), 
  Key Projects of National Natural Science Foundation of China (U1836222 and 61733011).}\\
\textsuperscript{\rm 1}Department of Computer Science and Engineering, Shanghai Jiao Tong University \\
\textsuperscript{\rm 2}Key Laboratory of Shanghai Education Commission for Intelligent Interaction \\ and Cognitive Engineering, Shanghai Jiao Tong University, Shanghai, China\\
\textsuperscript{\rm 3}MoE Key Lab of Artificial Intelligence, AI Institute, Shanghai Jiao Tong University, Shanghai, China\\
\{kingln,  felixxiao\}@sjtu.edu.cn, zhaohai@cs.sjtu.edu.cn
}
 
\begin{document}

\maketitle

\begin{abstract}
Named entity recognition (NER) models are typically based on the architecture of Bi-directional LSTM (BiLSTM). The constraints of sequential nature and the modeling of single input prevent the full utilization of global information from larger scope, not only in the entire sentence, but also in the entire document (dataset). In this paper, we address these two deficiencies and propose a model augmented with hierarchical contextualized representation: sentence-level representation and document-level representation. In sentence-level, we take different contributions of words in a single sentence into consideration to enhance the sentence representation learned from an independent BiLSTM via label embedding attention mechanism. In document-level, the key-value memory network is adopted to record the document-aware information for each unique word which is sensitive to similarity of context information. Our two-level hierarchical contextualized representations are fused with each input token embedding and corresponding hidden state of BiLSTM, respectively. The experimental results on three benchmark NER datasets (CoNLL-2003 and Ontonotes 5.0 English datasets, CoNLL-2002 Spanish dataset) show that we establish new state-of-the-art results.
\end{abstract}

\section{Introduction}

Named Entity Recognition (NER) is one of the fundamental
tasks in natural language processing (NLP) that intends to  identify words or phrases as the proper names of  PER (Person), ORG (Organization), LOC (Location), etc. 
Currently, most state-of-the-art NER systems \cite{huang2015bidirectional,lample2016neural,ma2016end,chiu2016named} employ BiRNNs, specially BiLSTM \cite{hochreiter1997long} as the encoder to extract the sequential information.

BiLSTM architectures exist limitations in making full use of global information. First, at each time step, BiLSTM takes current word embedding and past summary states as inputs, making it difficult to capture sentence-level information. 
\cite{zhang2018sentence} simultaneously model the sub-states for individual words and an overall sentence-level state.  \cite{liu2019gcdt} use a global contextual encoder and mean pooling strategy to capture sentence-level features, though they ignore the different importance of words in the same sentence.
Second, though BiLSTM updates the parameters with the iteration of all training instances, it only consumes one instance during both training and predicting. This nature prevents the model from effectively capturing document (dataset)-level information, e.g. for a unique token, its representations in training instances are indicative for recognizing the concerned token.
\cite{akbik2019pooled} use a pooling operation on different contextualized embeddings to generate global word representations. While they only consider the changes of embeddings for each unique word.

\begin{figure}[!t]
  \centering 
  \includegraphics[width=.998\columnwidth]{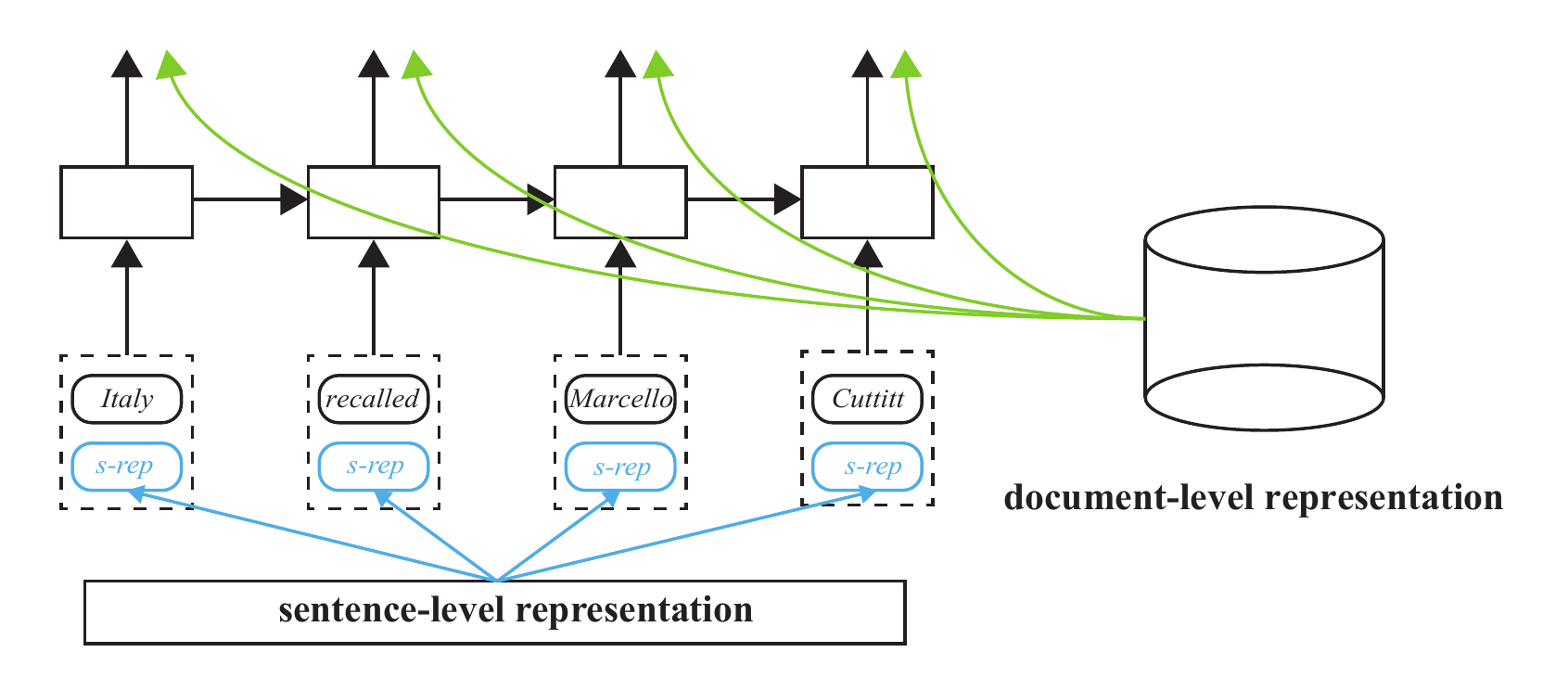}
    \caption{Incorporating hierarchical contextualized representation for NER. The sentence-level representation is assigned to each token and fed to the sequence labeling BiLSTM encoder. The document-level representation is fused with the hidden state of the BiLSTM and fed to the decoder.}\label{Overview}
\end{figure} 

In this paper, we propose a hierarchical contextualized representation architecture to enhance NER modeling. For sentence-level representation, inspired by \cite{wang2018joint}, we embed labels in the same space with word embeddings, and label embeddings are learned from attention mechanism computed with word embeddings to ensure that, each word embedding is much closer to their corresponding label embedding, and farther to other label embeddings. Then, the similarity between a word embedding and its nearest label embedding is regarded as a confidence score for this word. Hence, words with higher confidence scores contribute more to sentence-level representation. The sentence-level representations are then assigned to each token and fed to the encoder as shown in Figure \ref{Overview}. For document-level representation, we adopt a key-value memory component \cite{miller2016key} which memorizes all the word embeddings of training instances and their corresponding representations. The attention mechanism is adopted to compute the output of the memory component. The retrieved document-level representation is fused with the original hidden state and fed to the decoder as shown in Figure \ref{Overview}. In this case, the training instances are not only used to train the model parameters, but also involved in inference.

To verify the effectiveness our model, we conduct extensive experiments on three benchmark NER datasets.
Experimental results on these benchmarks suggest that our model can achieve state-of-the-art performance on CoNLL-2003 (91.96 $F_1$ without external knowledge and 93.37 $F_1$ with BERT), OntoNotes 5.0 (87.98 $F_1$ without external knowledge and 90.30 $F_1$ with BERT), 87.08 $F_1$ on CoNLL-2002, meaning that our model truly learns and benefits from useful contextualized representations.

Our contributions in this paper are summarized as follows.
\begin{itemize}
    \item We are the first to introduce hierarchical contextualized representations, namely sentence-level and document-level representation, for NER to take full advantage of non-local information.
    \item We introduce the label embedding attention mechanism for sentence-level representation and propose an effective approach to distill document-level information using key-value memory network.
    \item The evaluation results on three benchmark NER datasets show that our model outperforms all previously reported results without external knowledge. Furthermore, with pre-trained language model BERT, we establish new state-of-the-art results on CoNLL-2003 and ontonotes 5.0 datasets.
\end{itemize} 

\section{Related work}

\noindent\textbf{Neural Named Entity Recognition} 
Recently, with the development of deep neural network in a wide range of NLP tasks \cite{he2018syntax,he2019syntax,zhou2019head,xiao2019lattice,zhang2019dcmn+,zhang2019semantics}, neural network based models build 
reliable NER systems without hand-crafted features
or task-specific knowledge.
\cite{huang2015bidirectional} firstly proposed the BiLSTM-CRF architecture, which is used by most state-of-the-art models.
Later, character-level embeddings are concatenated to enhance the representation of rare and out-of-vocabulary words, these embeddings are generated with LSTM \cite{lample2016neural}, CNN \cite{ma2016end}, and recently IntNet \cite{xin2018learning}.
\cite{tran-etal-2017-named} stack BiLSTMs with residual connections between different layers of BiLSTM to add more representational
power. 
More recently, pre-trained language models from huge corpus are  adopted to enhance the representation of words \cite{peters2018deep,akbik2019pooled,devlin2018bert}. 

\noindent\textbf{Sentence-level Representation} 
 has been adopted to eliminate the limitations of RNNs due to their sequential nature. \cite{yang2017neural} leverage RNN models to learn sentence-level patterns for NER reranking.
\cite{zhang2018sentence} model the sub-states for individual
words and an overall sentence-level state simultaneously to
capture local and non-local contexts.
\cite{chen2019grn} use contextual layer and relation layer to model the relations between words in sentences, and then use gates to fuse local context features into global ones.
\cite{liu2019gcdt} simplify sentence-level state to average of the hidden states of each individual word from an independent global contextual encoder.
Inspired by \cite{wang2018joint} which use label information to construct text-sequence representations, we adopt label embedding attention to enhance the sentence-level representation learned from an independent BiLSTM.
The use of sentence-level information in \cite{liu2019gcdt} can be seen as a  special case of our model where the attention weight vector is a uniform distribution that assigns equal probabilities to all the words in the sentence.

\noindent\textbf{Document-level Representation}
\cite{qian2019graphie} considers the dependency structure of word sequence as the global information.
\cite{akbik2019pooled} dynamically aggregates contextualized embeddings for each unique string and then use a pooling operation
to generate a global word representation from these contextualized instances for NER. Different from their work which only uses the contextual word embeddings, our memory component uses the key-value memory networks to memorize the word representations (key) and the  hidden states (value) from the sequence labeling encoder. The attention mechanism is then called to calculate the document-level representation.

\section{Model}
\begin{figure*}[!t]
  \centering 
  \includegraphics[width=.98\textwidth]{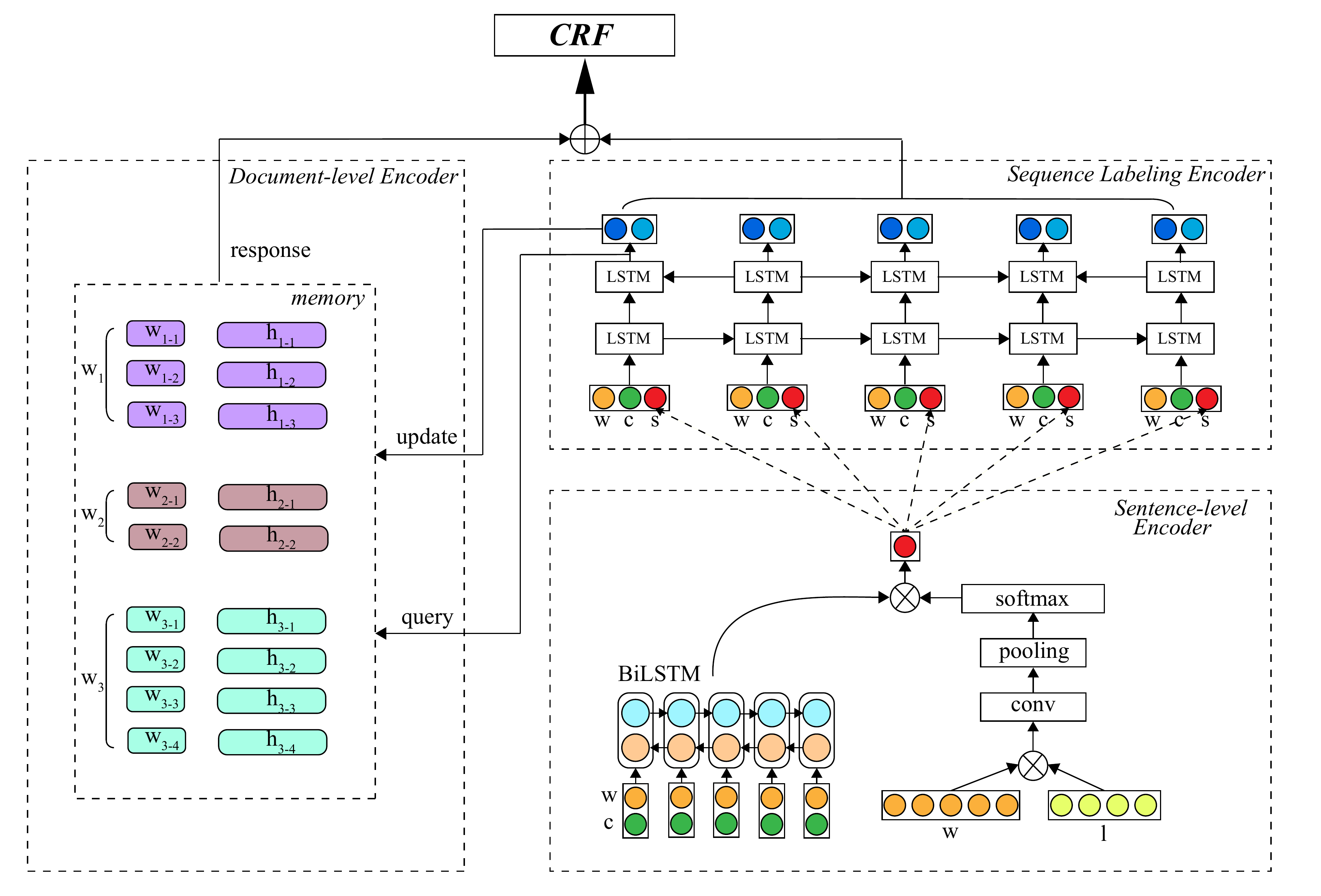}
    \caption{The main architecture of our NER model. The sequence labeling encoder (upper right) generates representations for the decoder and updates the memory component. The sentence-level encoder (bottom right part) generates the sentence-level representation, and the memory network (left part) computes the document-level contextualized information, in which the same color represents the memory slots for the unique word.}\label{Model}
\end{figure*}

This section presents our NER model in detail. The overall model architecture is shown in Figure \ref{Model}, which consists of four components: a decoder (top part), a sequence labeling encoder (upper right part), a sentence-level encoder (bottom right part), and a document-level encoder (left part). 

\subsection{Baseline Model}
We adopt the IntNet-LSTM-CRF model proposed by \cite{xin2018learning} as our baseline, which consists of three parts: representation module, sequence labeling encoder and decoder module.

\noindent\textbf{Token Representation}
Given a sequence of $N$ tokens $X= \{x_1, x_2, ..., x_N\}$, for each word $x_i$, we concatenate the word-level and character-level embedding as the joint word representation $x_i = [w_i; c_i]$. $w_i$ is the pre-trained word embedding. The character-level embedding $c_i$ is learned from IntNet, which
is a funnel-shaped wide convolutional neural architecture for learning representations of the internal structure of words. The network comprises of $L$ convolutions,  which implies $(L - 1) /2$ convolutional blocks. In each convolutional block, the first layer is the $N \times 1$ convolution which transforms the input, then the concatenation of convolutions with different kernel sizes in the second layer is fed to the next convolutional block. Direct connections from every other layer to all subsequent layers are used like dense connections. 

\noindent\textbf{Sequence Labeling Encoder}
The concatenation of word-level and character-level embeddings $x_i = [w_i; c_i]$  is then fed into the sequence labeling BiLSTM, which represents the sequential information at each step.

\begin{equation}
    \begin{aligned}
        h_i &= [\overrightarrow{h_i};\overleftarrow{h_i}] \\
        \overrightarrow{h_i} &= LSTM(x_i, \overrightarrow{h}_{i-1}; \overrightarrow{\theta}) \\
        \overleftarrow{h_i} &= LSTM(x_i, \overleftarrow{h}_{i-1}; \overleftarrow{\theta}) \\
    \end{aligned}
\end{equation}
where $\overrightarrow{\theta}$ and $\overleftarrow{\theta}$ are trainable parameters, respectively. 

\noindent\textbf{Decoder}
Conditional random field (CRF) \cite{Lafferty:01} has been widely used in state-of-the-art NER models \cite{lample2016neural,ma2016end} to help make decisions when considering strong connections between output tags.
During decoding, the Viterbi algorithm is applied to search the label sequence with the highest probability.
For $y=\{y_1,...,y_N\}$ being a predicted sequence of labels with same length as $x$. We define its score as:
\begin{equation}  
sc(x, y) = \sum_{i=0}^{N-1} Tr_{y_i, y_{i+1}} + \sum_{i=1}^{N} P_{i, y_i}
\end{equation}
where $Tr_{y_i, y_{i+1}}$ represents the transmission score  from the $y_i$ to $y_{i+1}$,
$ P_{i, y_i}$  is the score of the $j^{th}$ tag of the $i^{th}$ word from the sequence labeling encoder.

The CRF model defines a family of conditional probability $p(y|x)$ over all possible tag sequences $y$:
\begin{equation} 
p(y|x) = \frac{\exp^{sc(x,y)}}{\sum_{\tilde{y}\in y}\exp^{sc(x, \tilde{y})}} 
\end{equation}
during training, we consider the maximum log probability
of the correct sequence of tags. 
While decoding, we search the label sequence with maximum score:

\begin{equation}  
y^{*} = \arg\max_{\tilde{y} \in y} sc(x, \tilde{y})
\end{equation}

\subsection{Sentence-level Representation} 
Sentence-level  information has been shown highly
useful for model sequence \cite{zhang2018sentence,liu2019gcdt}.
We adopt an independent BiLSTM to generate contextualized features, which takes word representation $x_i=[w_i;c_i]$ as the input, we denote the hidden states of this BiLSTM as $v \in \mathbb{R}^{N \times d_s}$, where $N$ is the length of the sequence and $d_s$ is the hidden size. 
Considering that words may contribute differently to the sentence-level representation, we adopt label embedding attention \cite{wang2018joint} to get an attention score for the entire sentence and then transform the hidden states $v \in \mathbb{R}^{N \times d_s}$ into a fixed-sized sentence-level representation $s \in \mathbb{R} ^{d_s} $.  

As shown in Figure \ref{Model}, we embed all the label types (e.g. LOC, PER, etc.) in the same space as the word embeddings. We denote the label embeddings as $l = [l_1, l_2, ... l_P], l \in \mathbb{R}^{P \times d_w} $, where  $P$ is the number of labels, $d_w$ is the dimension of the word embeddings. 
We train on the training instances to ensure that, each word embedding is more closer to their corresponding label embedding, and farther to other label embeddings. 
For example, the token $\emph{Italy}$ is labeled as LOC type in the example of Figure \ref{Overview}, we attempt to make it closer to the label embedding of LOC, and farther to other label embeddings (e.g. the label embedding of PER).
The cosine similarity
\footnote{The reason for using cosine similarity is the same as the next subsection, and will be analyzed later.} 
$e(x_i, l_j)$ between the word embeddings $x_i$ and the label embedding $l_j$ can be taken to measure the confidence score of this word-label pair:

\begin{equation}
    e(x_i, l_j) = \frac{x_i^Tl_j}{\|x_i\|\|l_j \|}
    % e_i(l_j) = \frac{x_i^Tl_j}{\|x_i\|\|l_j\|}
\end{equation}

We use convolutional neural network (CNN) to capture the relative spatial information among consecutive words in the sentence. Further, the largest confidence score $m_i  \in \mathbb{R} ^{P} $ between the $i$-th word and all labels is obtained by a max-pooling operation:

\begin{equation}
    m_i = max(W^T\left[
    \begin{array}{c}
      e(i - \frac{k-1}{2}, :) \\
      \cdots \\
     e(i + \frac{k-1}{2}, :)
    \end{array}
\right] + b)
\end{equation} 
where $W \in \mathbb{R} ^{k} $ and $b \in \mathbb{R} ^{P}$ are trainable parameters, $k$ is the kernel size, $max$ denotes max pooling.

The attention (confidence) score $\beta \in \mathbb{R} ^{N} $ for the entire sentence is:

\begin{equation}
    % \beta = \frac{e^{m_i}}{\sum_{j=1}^Ne^{m_j}}
    \beta =  softmax(m)
\end{equation}

The sentence-level representation $s \in \mathbb{R} ^{d_s}$ can be simply obtained via averaging the hidden states $v \in \mathbb{R} ^{N \times d_s}$, weighted by the attention score calculated above:

\begin{equation}
    s = \sum_{i=1}^N\beta_iv_i
\end{equation}

The sentence-level representation $s \in \mathbb{R} ^{d_s} $ is then concatenated with the word representation $x_i' = [x_i; s]$ and fed to the sequence labeling encoder.
Note that the training and test process are the same. We use the label embeddings $l \in \mathbb{R}^{P \times d_w}$ of all labels, not the ground-truth label embedding.
The intuition for using the label embedding attention is that each word in the sentence contributes differently to sentence-aware representation. The similarity between each word embedding and its nearest label embedding can be regarded as the confidence score of this word-label pair. Words with higher confidence score should contribute more to the sentence-level contextualized representation.

\subsection{Document-level Representation} 

In terms of memory network, we introduce document-aware representations of the unique word in training instances as an extra knowledge source to help the prediction.  Memory network was originally proposed by \cite{weston2014memory} in the domain of question answering (QA) for prediction, where the long-term memory acts as a dynamic knowledge base.
\cite{miller2016key} further introduces a key-value memory networks, which utilize different encodings in the addressing and output stages. The keys are designed to help match the question, while the values are to generate the response. 

We adopt the key-value memory component $\textit{M}$  to memorize document-level contextualized representation. 
Memory slots are defined as pairs of vectors $(k_1, v_1),. . . ,(k_m, v_m)$.
% and denote the query word as $q$.
In each single slot, the key represents the word embedding $w_i$, and the value is the corresponding hidden states $h_i$ from sequence labeling encoder for each token in training instances. So the same word may occupy in many different slots because of changing embeddings and representations under different contexts.
Table \ref{example} shows an example of using training instances to help indicate the NE type of queried token.

\noindent\textbf{Memory Update} 
The word embeddings are fine-tuned during training and used to update the key part of the memory. The sequence labeling encoder generates the hidden states to update the value part. Supposing the states of the $i$-th token is changed after computation, the $i$-th slot in the memory $M$ will be rewritten.
Each memory slot will be updated once in one epoch.

\noindent\textbf{Memory Query} 
For the $i$-th word in the sentence, we distill all the contextualized representations for this word in the memory $M$ through an inverted index that finds a subset $(k_{sub_1},v_{sub_1}), ..., (k_{sub_T}, v_{sub_T})$ of size $T$, where the inverted index records the positions of the unique word in the memory $M$ as shown in Table \ref{example}. $T$ represents the number of occurrences of this word among the training instances.

The attention operation is called to compute the weight of document-level representation. For the unique word, the memory key $k_j \in [k_{sub_1}; ...; k_{sub_T}]$ is used as the attention key, the memory value $v_j \in [v_{sub_1}; ...; v_{sub_T}]$ is used as the attention value. Then the embedding $w_{q_i}$of the queried word serves as the attention query $q_i$. Here, we consider three compatibility functions $u_{ij} = o(q_i, k_j)$:

\begin{table}[t!] 
\centering
\resizebox{.998\columnwidth}{!}{
\begin{tabular}{l}
\toprule
\textbf{Test instance} \\ 
\midrule 
\emph{\textbf{Italy} recalled Marcello Cuttitt.} \\
\midrule\midrule
\textbf{Training instances} \\
\midrule 
1. \emph{ORVIETO} (0), \textbf{\emph{Italy} (1)} \emph{1996-08-24} (2).  \\
2. \emph{Rohrabacher} (3) \emph{had} (4) \emph{recently} (5) \emph{visited} (6) \textbf{\emph{Italy} (7)} \\
3. \emph{Andrea} (8) \emph{Ferrigato} (9) \emph{of} (10) \textbf{\emph{Italy} (11)} \emph{sprinted} (12)...\\
\midrule\midrule
\textbf{Inverted index} \\
\midrule
{\emph{Italy}: [1, 7, 11, ...]} \\
\bottomrule
\end{tabular}}
\caption{Query operation for the word \emph{Italy}.  The numbers in parentheses indicate slot index of tokens in memory $M$. The memory slots of these bold tokens in  training instances are retrieved according to the inverted index for \emph{Italy}. 
}\label{example}
\end{table}

\noindent{(1) dot-product attention}
\begin{equation}
    o_1(q_i, k_j) = q_ik_j^T
\end{equation}
(2) scaled dot-product attention \cite{vaswani2017attention}
\begin{equation}
    o_2(q_i, k_j) = \frac{q_ik_j^T}{\sqrt{d_w}}
\end{equation}
and (3) cosine similarity 
\begin{equation}
    o_3(q_i, k_j) = \frac{q_ik_j^T }{\|q_i\|\|k_j\|}
\end{equation}
where $d_w$ represents the dimension of word embeddings.

\noindent\textbf{Memory Response}
The document-level representation is computed as:

\begin{equation}
    \begin{aligned} 
        \alpha_{ij} & = \frac{\exp(u_{ij})}{\sum_{z=1}^T\exp(u_{iz})} \\
        r_i&= \sum_{j=1}^T\alpha_{ij}v_j
    \end{aligned} 
\end{equation}

Then the fusion representation $g_i \in \mathbb{R} ^ { d_h}$ of the original hidden representation and this document-level representation are fed to the CRF layer, where $d_h$ is the hidden size of the sequence labeling encoder.
\begin{equation}
    g_i = \lambda h_i + (1 - \lambda)r_i
\end{equation}
where $\lambda$ is a hyperparameter, indicating how much document-aware information is adopted, 0 for document-level representation only and 1 for discarding all document-level information at all.

\section{Experiment}

\subsection{Dataset}
Our proposed representations are evaluated on three benchmark NER datasets: CoNLL-2003 \cite{sang2003introduction} and OntoNotes 5.0  \cite{pradhan2013towards} English NER datasets, CoNLL-2002 Spanish NER \cite{Erik2002introduction} dataset. 

\begin{itemize}
    \item \textbf{CoNLL-2003 English NER} consists of 22,137 sentences
    totally and is split into 14,987, 3,466 and 3,684 sentences
    for the training, development set and test sets, respectively. It is tagged with four linguistic
    entity types (PER, LOC, ORG, MISC).
    \item  \textbf{CoNLL-2002 Spanish NER} consists of 11,752 sentences totally and is split into 8,322, 1,914 and 1,516 sentences
    for the training, development and test sets, respectively. It is also tagged with four linguistic
    entity types (PER, LOC, ORG, MISC).
    \item \textbf{OntoNotes 5.0} consists of 76,714 sentences
    from a wide variety of sources (magazine, telephone conversation,
    newswire, etc.). Following \cite{chiu2016named,chen2019grn}, we use the portion of the dataset with gold-standard named
    entity annotations, and thus exclude the New Testaments
    portion. It is tagged with eighteen entity types (PERSON, CARDINAL, LOC, PRODUCT, etc.).
\end{itemize}

\begin{table}[t!] 
\centering
\resizebox{.998\columnwidth}{!}{
\begin{tabular}{l|c}
\hline
\textbf{\textbf{Models}} & \textbf{$F_1$}\\
\hline 
% \cite{zhuo2016segment} & 88.12 \\
\cite{lample2016neural} & 90.94 \\
\cite{ma2016end} & 91.21 \\
% \cite{rei2017semi} & 86.26 \\
\cite{yang2017neural} & 91.62 \\
\cite{liu2018empower} & 91.24 $\pm$ 0.12 \\
% \cite{ye2018hybrid} & 91.38 $\pm$ 0.10 \\
\cite{yang2018ncrf} & 91.35 \\
\cite{zhang2018sentence} & 91.57 \\ 
\cite{xin2018learning} & 91.64 $\pm$ 0.17 \\  
\cite{liu2019contextualized} & 91.10 \\
% \cite{liu2019learning} & 91.47 \\
\cite{chen2019grn} & 91.44 $\pm$ 0.10 \\
\cite{qian2019graphie} & 91.74\\
\cite{liu2019gcdt} \footnotemark[2] & 91.54 \\
\hline
Ours & \textbf{91.96 $\pm$ 0.03} \\
\hline\hline
\multicolumn{2}{l}{\textbf{+ Language Models / External knowledge}} \\ 
\hline
% \cite{collobert2011natural} & 89.59 \\
% \cite{luo2015joint} & 91.20 \\
\cite{chiu2016named}$\dagger$ & 91.62 $\pm$ 0.33 \\
\cite{liu2018empower} & 91.71 $\pm$ 0.10\\ 
\cite{peters2018deep} (ELMo) & 92.20 \\
\cite{clark2018semi} & 92.61\\ 
\cite{devlin2018bert} (BERT)& 92.80\\
\cite{akbik2018contextual}$\dagger$ &93.09\\ 
% \cite{guo-etal-2019-star} & 91.98 \\
\cite{akbik2019pooled}$\dagger$ & 93.18\\
\cite{liu2019gcdt} (BERT) \footnotemark[2] & 93.23 \\
\hline 
Ours + BERT & \textbf{93.37 $\pm$ 0.04} \\
\hline
\end{tabular}}
\caption{$F_1$ scores on CoNLL-2003. 
% Scores for our methods
% are the average of 5 runs. 
$\dagger$ refers to models
trained on both training and development datasets.}
\label{03reult}
\end{table} 

 \footnotetext[2]{Through personal communication, the authors confirmed that they directly tested the BIOES tagged results with the official conlleval script (which can only works for BIO tagged entities), giving the results reported in their paper 91.96 / 93.47, while our re-evaluation results are 91.54 / 93.23 with strict BIO tag converting from the released file by the authors.}

\noindent\textbf{Metric}  We use the BIOES sequence labeling
scheme instead of BIO for these three datasets during training. As for test, we convert the prediction results back to the
BIO scheme and use the standard conlleval script to compute the $F_1$ score.

\subsection{Setup} 
\textbf{Pre-trained Word Embeddings.} For the CoNLL-2003 and OntoNotes 5.0 English datasets, we use the publicly available pre-trained 100$D$ GloVe \cite{pennington2014glove} embeddings. For CoNLL-2002 Spanish dataset, we train 64$D$ GloVe embeddings with the minimum frequency of occurrence as 3, and the window size of 5. The word embeddings are fine-tuned during training.\\
\textbf{Character Embeddings.} We train the IntNet character embeddings \cite{xin2018learning}. The dimension of character embeddings is 32, which is randomly initialized, the filter size of the initial convolution is 32 and that of other convolutions is 16. Different from \cite{xin2018learning}, we set filters as size [3; 5] for all the kernels, and the number of convolutional layers is 7.
\\
\textbf{Parameters.}  We follow the work of \cite{yang2018ncrf}, and conduct optimization with the stochastic gradient descent \footnote{Code will be available at https://github.com/cslydia/Hire-NER.}. The batch size is set as 10, the initial learning rate is set to 0.015 and will shrunk by 5\% after each epoch. The hidden size of sequence labeling encoder and the sentence-level encoder are set as 256 and 128, respectively. We apply dropout 
% \cite{srivastava2014dropout}
to embeddings and hidden states with a rate of 0.5. The $\lambda$ used to fuse original hidden state and document-level representation is set as 0.3 empirically. 
For each type of NEs, we randomly select hundreds of NEs, and calculate the average of the word embeddings as its label embedding. 

\begin{table}[t!] 
\centering
\resizebox{.998\columnwidth}{!}{
\begin{tabular}{p{5.6cm}|c}
\hline
\textbf{\textbf{Models}} & \textbf{$F_1$}\\
\hline 
\cite{gillick2015multilingual} & 82.95 \\
\cite{lample2016neural} & 85.75 \\
\cite{yang2017transfer} & 85.77 \\ 
\cite{xin2018learning} & 86.68 $\pm$ 0.35 \\ 
\hline
Ours & \textbf{87.08 $\pm$ 0.16} \\
\hline
\end{tabular}}
\caption{$F_1$ scores on CoNLL-2002.}
\label{02result}
\end{table} 

\begin{table}[!t] 
\centering
\resizebox{.998\columnwidth}{!}{
\begin{tabular}{p{5.9cm}|c}
\hline
\textbf{\textbf{Models}} & \textbf{$F_1$}\\
\hline 
\cite{durrett2014joint} & 84.04\\
\cite{chiu2016named} & 86.28 $\pm$ 0.26 \\
\cite{shen2017deep} & 86.63 $\pm$ 0.49 \\
\cite{strubell2017fast} & 86.84 $\pm$ 0.19 \\
\cite{ghaddar2018robust} & 87.44 \\
\cite{chen2019grn} & 87.67 $\pm$ 0.17 \\ 
\hline
Ours & \textbf{87.98 $\pm$ 0.05} \\
\hline\hline 
\multicolumn{2}{l}{\textbf{+ Language Models / External knowledge}} \\ 
\hline
\cite{ghaddar2018robust} & 87.95 \\ 
\cite{clark2018semi} & 88.88 \\
\cite{akbik2019pooled}\footnotemark[3] & 89.71\\ 
\hline
Ours + BERT & \textbf{90.30}\\
\hline
\end{tabular}}
\caption{$F_1$ scores on OntoNotes 5.0.}
\label{ontoresult}
\end{table}  

\footnotetext[3]{The authors of  \cite{akbik2019pooled} released the result of 89.3 in their github \url{https://github.com/zalandoresearch/flair}, 89.71 is our re-implement result.}

\subsection{Results and Comparisons} 
Tables \ref{03reult}, \ref{02result}, \ref{ontoresult} compare our model to existing state-of-the-art approaches on the three benchmark datasets. Our model surpasses previous state-of-the-art approaches on all the three datasets.
On CoNLL-2003 dataset, we compare our model with the state-of-the-art models, including the  models that use global information to enhance the representation \cite{yang2017neural,zhang2018sentence,qian2019graphie,liu2019gcdt}. 
We also incorporate pre-trained language model BERT \cite{devlin2018bert} for fair comparisons with the models which also use pre-trained language models or other external knowledge.
Some of the results \cite{akbik2018contextual,akbik2018contextual} are not comparable to our results directly, because their final models are trained on both training and development datasets.
On CoNLL-2002 Spanish dataset, our model achieves 87.08 $F_1$ score without external knowledge, which surpasses previous best score by 0.4.
Considering that the above two datasets are relatively small, we further conduct experiment on a much more large OntoNotes 5.0 dataset, which also has more entity types. 
We compare our model with the previous model that also reported results on it \cite{chiu2016named,shen2017deep,ghaddar2018robust}.
 As shown in Table \ref{ontoresult}, our model shows a significant advantage on this dataset, which outperforms previous state-of-the-art results substantially at 87.08 (+0.31) without BERT,  and 90.30 (+0.59) with BERT. More notably, our model without external knowledge surpasses the previous model \cite{ghaddar2018robust}, which use extra lexicon information of 120 entity types from Wikipedia.
 Overall, the comparisons on these three benchmark datasets well demonstrate that our model truly learns and benefits from useful sentence-level and document-level representation without the support from external knowledge.
 
 \begin{table}[t!] 
\centering
\resizebox{.998\columnwidth}{!}{
\begin{tabular}{lccc}
\toprule
\textbf{\textbf{}} & CoNLL03 & CoNLL02 & OntoNotes\\
\midrule 
base model & 91.60 & 86.65 & 87.58\\
 + sentence-level & 91.80 & 86.95 & 87.86 \\
+ document-level & 91.79 & 86.76 & 87.81 \\ 
+ ALL & \textbf{91.96} & \textbf{87.08} & \textbf{87.98} \\ 
\bottomrule
\end{tabular}}
\caption{Ablation study on the three benchmark datasets.}
\label{ablation}
\end{table}  

\begin{table}[t!] 
\centering
\resizebox{.998\columnwidth}{!}{
\begin{tabular}{|l|c|p{1cm}|p{1cm}|}
\hline
& Strategy & $F_1$ & ERR\\
\hline 
base model & - & 91.60 & - \\
\hline\hline
\multirow{2}{*}{sentence-level} & mean-pooling & 91.65 & 0.60 \\
\cline{2-4}
& label-embedding & 91.80 & 2.23 \\
\hline\hline
\multirow{3}{*}{document-level} & dot-product & 91.63 & 1.55 \\
\cline{2-4}
& scaled dot-product & 91.75 & 1.79 \\
\cline{2-4}
& cosine similarity & 91.79 & 2.38 \\
\hline\hline
ALL & - & \textbf{91.96} & \textbf{3.81} \\ 
\hline
\end{tabular}}
\caption{Comparison of different strategies on CoNLL-2003 dataset. ERR is the relative error rate reduction of our model compared to the baseline.}
\label{strategy}
\end{table} 

 \subsection{Ablation Study}
 In this experiment, we individually adopt two hierarchical contextualized representations to enhance the representation of tokens: sentence-level representation for assigning the sentence state to each token and document-level representation for inference. Table \ref{ablation} shows the $F_1$ score raise and relative error reduction brought by each of the two hierarchical representation on the three benchmark datasets. We discover that both sentence-level and document-level representations enhance the baseline. By combing these two representations together, we get a larger gain of 0.36 / 0.43 / 0.40, respectively.
 
We further analyze the two hierarchical representations by adopting different strategies. \cite{liu2019gcdt} perform mean pooling over all the tokens to generate sentence-level representation. We further conduct experiments to investigate the three compatibility functions used to employ memorized information. As shown in Table \ref{strategy}, compared with the mean pooling strategy, our label-embedding attention mechanism raises the $F_1$ score by 0.25. Among the three compatibility functions to compute the weight of query word and memorized slots,
cosine similarity performs best, while dot-product performs worst.
\cite{vaswani2017attention} use scaled dot-product to counteract the dot products growth in magnitude, showing better than dot product.
Cosine similarity calculates the inner product of word vectors with unit length, and can further solve the inconsistency between the embeddings and the similarity measurement. Thus, we eventually adopt cosine similarity as the compatibility function.

\begin{table}[t!] 
\centering
\resizebox{.998\columnwidth}{!}{
\begin{tabular}{cccccccc}
\toprule
& \multicolumn{3}{c}{\textbf{Baseline}} & & \multicolumn{3}{c} {\textbf{Ours}}\\ 
\cmidrule{2-4} \cmidrule{6-8}
\textbf{} & {$P$} & {$R$}  & {$F_1$} &  &{$P$} & {$R$}  & {$F_1$} \\
\midrule 
IV &  94.58 & 93.16 & 93.87 && 94.96 & 93.58 & \textbf{94.26}\\
OOTV & 93.46&  91.57 & 92.51 && 94.07 &  91.85 &\textbf{92.95} \\
OOEV & 94.12 & 94.12 & 94.12 && 94.12 & 94.12 & 94.12 \\ 
OOBV &88.42 & 84.81 & 86.58 & & 88.51 &  85.56 & \textbf{87.01}  \\
\bottomrule
\end{tabular}}
\caption{Detailed results on the CoNLL-2003 dataset for IV, OOTV, OOEV, OOBV. }
\label{detail}
\end{table}

\subsection{Memory Size and Time Consuming} 
\begin{figure}[!t]
  \centering 
  \includegraphics[width=.998\columnwidth]{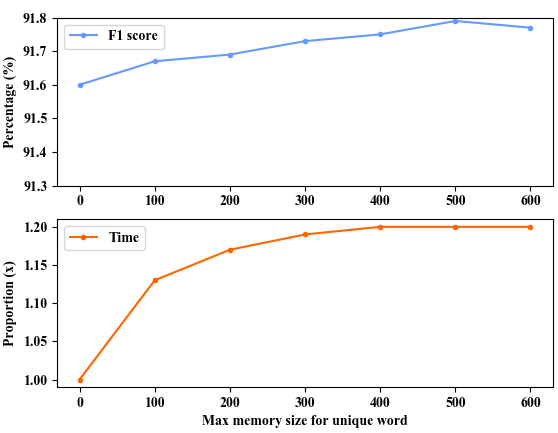}
    \caption{$F_1$ score and time proportion with respect to the max memory size. Time proportion represents the ratio of our training time compared to baseline. }\label{memorysize}
\end{figure}

 Figure \ref{memorysize} illustrates our model performance and time proportion compared to the baseline with respect to the max queried subset size $T$ for each unique word in the memory query step. For words occurring more than $T$ times in the corpus (these words are more likely to be stop words when $T$ is large), we only randomly select $T$ slots in the subset to compute the  document-level representation.
For fair comparisons, we keep the IntNet layer, sequence labeling layer and CRF layer the same for all the experiments. The consumed time of our model is only 19\% more than the baseline  on CoNLL-2003 dataset even with the max memory size as 500.
Therefore, our model  brings slight increase on time consumption. 
When $T$ is less than 500, the larger $T$ may incorporate more useful contextualized representation for practice words and improve the results accordingly, when $T$ is larger than 500, which may involve more stop words, our model drops slightly.

\subsection{Improvement Discussion}
 Table \ref{detail} presents the $F_1$ score of in-both-vocabulary words (IV), out-of-training-vocabulary words (OOTV), out-of-embedding-vocabulary words (OOEV), and out-of-both-vocabulary words (OOBV) on CoNLL-2003 datset. According to our statistic, 63.40\% / 52.43\% / 84.68\% of the NEs in the test set of CoNLL-2003, CoNLL-2002, and OntoNotes datasets are located in the IV part, respectively. Therefore, it is of great importance to focus on this part. We adopt memory network to memorize and retrieve the global representations and use the memorized training instances directly to participate in inference, which greatly improves both the precision and recall of the NEs in IV part, in which our model outperforms baseline by 0.39 in terms of $F_1$ score.
 For OOV NEs, sentence-level representation can help these concerned tokens aware of the entire sentence, thus enhance the performance. The improvement is 0.44 / 0.43 $F1$ score for OOTV NEs and OOBV NEs, respectively.

\section{Conclusions} 
In this paper, we adopt hierarchical contextualized representations to enhance the performance of named entity recognition (NER). Our model makes full use of the training instances and the spatial information of the embedding space  by incorporating sentence-level representation and document-level representation.
We consider the importance of words in the sentences and weight their contributions with the label embedding attention for the sentence-level representation. 
For words shown  in training instances, we memorize the  representations of these instances, and involve these representations for inference during test. 
Empirical results on three benchmark datasets (CoNLL-2003  and Ontonotes 5.0 English datasets, CoNLL-2002 Spanish dataset)
show that our model outperforms previous state-of-the-art systems with or without pre-trained language models respectively.

\bibliography{AAAI-LuoY.7327}
\bibliographystyle{aaai}

\end{document}